\documentclass[journal]{IEEEtran}
\usepackage{multirow}
\usepackage{amsthm,amscd,algorithm,algpseudocode}
\usepackage{graphicx} % inserting images
\DeclareGraphicsExtensions{.jpg,pdf,.png}
\ifCLASSINFOpdf
\else
\fi

\pdfoutput=1
\title{Using Empirical Covariance Matrix in Enhancing Prediction Accuracy of Linear Models with Missing Information}

\author{
	 Ahmadreza Moradipari, Sina Shahsavari, Ashkan~Esmaeili
	and~Farokh~Marvasti\\Advanced Communications Research Institute (ACRI), and\\Electrical Engineering Department, Sharif University of Technology, Tehran, Iran\\aesmaili@stanford.edu, moradipari@ee.sharif.edu% <-this % stops a space
	\thanks{Moradipari, Shahsavari, and Marvasti are with the Advanced Communications Research Institute (ACRI), and Department of Electrical Engineering, Sharif University of Technology. Ashkan Esmaeili was with Stanford Electrical Engineering Department, and is Stanford MS alumni. He is now on Ph.D program with Electrical Engineering Department at Sharif University of technology and ACRI.}% <-this % stops a space
	}
\begin{document}
\maketitle
\begin{abstract}
~Inference and Estimation in Missing Information (MI) scenarios are important topics in Statistical Learning Theory and Machine Learning (ML). In ML literature, attempts have been made to enhance prediction through precise feature selection methods. In sparse linear models, LASSO is well-known in extracting the desired support of the signal and resisting against noisy systems. When sparse models are also suffering from MI, the sparse recovery and inference of the missing models are taken into account simultaneously. In this paper, we will introduce an approach which enjoys sparse regression and covariance matrix estimation to improve matrix completion accuracy, and as a result enhancing feature selection preciseness which leads to reduction in prediction Mean Squared Error (MSE). We will compare the effect of employing covariance matrix in enhancing estimation accuracy to the case it is not used in feature selection. Simulations show the improvement in the performance as compared to the case where the covariance matrix estimation is not used.
\end{abstract}
\section{Intorduction}
Recently, inference and learning in problems which are suffering from incomplete datasets have gained specific attention since these types of problems are accompanied with important applications in practical settings. In \cite{esmaeili2016fast}, practical settings are introduced where dealing with missing information and developing a statistical model which could learn the incomplete data are necessary. It is intuitively comprehensible that learning procedures would differ in missing data scenarios. We illustrate a novel example which could clarify what dealing with missing data means. Consider an image has missing information or lossy segments, and as a result, many pixels are lost or corrupted. In this setting, knowing the fact that the image is low-rank (few colors are used in painting for instance) helps restoration. As another illustration, suppose we have many athletes for whom specific records and measurements are gathered. The recorded features may have lots of features as not reported or not assigned (NA). Our purpose may be deriving the statistical model which determines the weight of each feature affecting a specific criterion such as athlete's stamina. Thus, we have a learning problem in a missing information scenario. Many works in the literature have been carried out towards the matrix completion problem including but not limited to \cite{5466511}, and \cite{candes2009exact}.
Now, we proceed to add another aspect into the problem. In many settings, we are actually solving Compressed Sensing (CS) problems \cite{candes2006compressive},  where there is some sparsity pattern in the problem model. We briefly review how CS became popular in the literature. In many problems, there exists some sparsity pattern which leads to finding a unique solution for underdetermined settings provided there are sparsity constraints. CS problems have many well-known methods developed in the literature which could be classified in three main classes. One class is related to the soft-thresholding methods. The most famous soft thresholding method which approximates $l_0$- norm with $l_1$-norm is Lasso. The other class consists of greedy methods known as hard thresholding methods such as Iterative Hard Thresholding (IHT) \cite{tropp2007signal}, Iterative Method with Adaptive Thresholding (IMAT) \cite{marvasti2012unified}, Smoothed L0 (SL0) \cite{4663911}, and Iterative Null-Space Projection Method of Adaptive Thresholding (INPMAT) \cite{esmaeili2016iterative}. Interestingly, we observe that in the practical settings we usually confine the large incomplete datasets to supporting columns and believe that in the oracle model, the desired variable is regressed on very few (limited) number of features. For example, in the athlete case, a few features affect the stamina, and as a result the parameters (weights) vector in regression is assumed to be sparse. Consequently, we are dealing with a model which learns in missing information settings knowing that some sparsity pattern exists in the model. In fact, this could be considered as a joint problem of missing data imputation and sparse recovery of the parameters vector (supporting columns). Similar setting for matrices is investigated in \cite{doi:10.1137/100781894} by Tao, et al. \\
~~~~The rest of the paper is organized as follows: In Section II, we explain our algorithm and compare it to that in \cite{esmaeili2016fast}. In Section III, we provide the simulation results and discuss the efficacy of out method in enhancing prediction accuracy. Finally, Section IV, concludes the paper.
\section{Methodology}
The authors in \cite{esmaeili2016fast} introduced a new fast approach in dealing with the combined problem of matrix completion and sparse recovery. In \cite{esmaeili2016fast}, authors mention that precise matrix completion and data inference on large structures are based on algorithms which are computationally complex. The well-known algorithms for matrix completion in the literature are based on singular value decomposition in consecutive iterations which does not seem to be reasonable for big data scenarios as stated in \cite{esmaeili2016fast}. Therefore, the authors claim that it is sufficient to implement the completion algorithm for few iterations or even apply a simple (non-complex) completion method which leads to a good approximation of the support, and afterwards confine the data matrix on the approximated support. The next step in their proposed algorithm is to apply the complex and precise completion methods on the shrunken size data matrix so as to enhance prediction accuracy as well as reducing runtime. They also added a leg-up to their method by introducing augmented version of their proposed method which enhances the prediction accuracy. As the simulations show in that paper, the complexity is reduced significantly; however the prediction accuracy is worse than the scenario in which the data matrix is pre-completed precisely.
In this paper, our aim is to compensate for the accuracy by employing the empirical covariance matrix estimation. It is worth noting that the algorithm presented in \cite{esmaeili2016fast} and the one we are going to mention are applicable to every method in sparse linear regression with missing data. For example, two methods on this specific topic are presented in \cite{ganti2015sparse}, \cite{chan2016temporal}, \cite{goldberg2010transduction}.
We explain the steps used in our algorithm as follows: First, we pre-complete the data matrix with a simple inexact method. Then, after we continued the completion until the support achieved by LASSO is a viable support, we form the empirical covariance matrix for the features in the support. In the next step, we choose the variables which have high correlation with chosen variables in the support. Next, we confine the initial data matrix with missing information to the columns whose indices come from the features opted from the LASSO solution nurtured with the indices of variables which are highly correlated to them. Now, we apply a precise completion method on the selected columns of the matrix. The reason we proceed this way is that in comparison to the algorithm stated in \cite{esmaeili2016fast}, although we have slightly more complexity, we enhance the prediction accuracy. The reason for increase in the prediction accuracy is that the added variables to the chosen support help reconstruct the structure of sensing matrix. Although they may not affect the sparse regression variables chosen in the support of parameters vector, they help achieve better matrix completion than only considering the support variables as the columns considered for completion as stated in \cite{esmaeili2016fast}.
\section{Problem Model}
The linear mode of interest is as follows:
\begin{equation}
y=X\beta+\epsilon,
\end{equation}
where $y$ is the measurement vector, $X$ is the data (sensing) matrix with missing information, $\beta$ is the sparse weights (parameters) vector, and $\epsilon$ is the noise in the model. The underlying assumptions on this linear model could be best explained as a problem which is simultaneously dealing with missing information and sparse recovery. Previously, the scenarios in which we have fully known the data matrix as appeared in compressed sensing and regression problems differ from our case. Generally, there is no guarantee that we fully know the data matrix. In practice, we have several cases where this could happen. The data matrix entries could be missing due to costs in measurements, lacking access to report some data, inaccurate measurement device which leads to invalid report, and as a result, validation of the measured data is questioned. There are numerous practical settings suffering from the missing (corrupted) information such as massive MIMO, recommendation systems, medical datasets, etc. Thus, the problem is a generalization of the sparse recovery problem to a combined problem which has to recover the sparse parameters as well as inferring and completing the structure of data matrix. 
\begin{algorithm}
		\caption{Four-step Recovery}\label{tab:L2SAT}
		\begin{algorithmic}[1]
			\State \textbf{Input:}
			\State  $\mathbf{Y_{m\times1}}: $ {The vector containing the labels}
			\State  $\mathbf{{\hat{X}_{m\times n}}}: $ {The data matrix containing missing entries}
	
			\State  $\mathbf{\epsilon}: $ {Stopping criterion}
			\State  $(\mathbf{\alpha},\mathbf{\lambda_1},\mathbf{\lambda_2})$ {Algorithm Parameters}
			\State \textbf{Output:}
		\State  $\beta^*$ : {The reconstructed signal}
			
			\Procedure{Four-step Recovery} {${\mathbf Y},{\mathbf {\hat{X}}},{\epsilon},{\alpha}, \lambda_1, \lambda_2$}
\State {Initialization:} $\mathbf{X_0}\gets \mathbf{0}_{m\times n}, k\gets 0$
		\State  $\mathbf{\beta_{0_{n\times1}}} \gets \mathbf{0}_{n\times 1},\mathbf{\beta_{1_{n\times1}}} \gets \mathbf{\hat{X}^{\dagger}} \times \mathbf{Y}$
		    \While {$||\beta_k-\beta_{k+1}||_2>\epsilon$}
 			\State {fix $\beta_k$ and solve the following: }
 			\State {$\mathbf{X_k} \gets \min_{X_k} ~ ||P_E(\mathbf{X_k}-\mathbf{\hat{X}})||_2^2+\lambda_1||\mathbf{X_k}||_*$}
			\State {fix $\mathbf{X}_k$ and solve the following: }			
 			\State {$\beta_{k+1}^*(\lambda_2) \gets \min_{\beta} ~ ||\mathbf{X_k}\beta-Y||_2^2+\lambda_2||\beta||_1$}
			\State {$k \gets k+1$ }			
			\EndWhile
			 			\State {$S \gets \{i: \beta_i \neq 0\}$, $s \gets |S|$ }
			\State {Confine $\mathbf{\hat{X}}$ on supporting columns in $S$, and denote it with $\mathbf{\hat{X}_{S}}$. Choose $\alpha<<\epsilon$.}
			\State {Initialization:} $\mathbf{X_0}\gets \mathbf{0}_{m\times s},  k\gets 0$
					\State  $\mathbf{\beta_{0_{n\times1}}} \gets \mathbf{0}_{s\times 1},\mathbf{\beta_{1_{n\times1}}} \gets \mathbf{\hat{X}_S^{\dagger}} \times \mathbf{Y}$
		    \While {$||\beta_k-\beta_{k+1}||_2>\alpha$}
 			\State {fix $\beta_k$ and solve the following: }
 			\State {$\mathbf{X_k} \gets \min_{X_k} ~ ||P_E(\mathbf{X_k}-\mathbf{\hat{X}_{{S}}})||_2^2+\lambda_1||\mathbf{X_k}||_*$}
			\State {fix $\mathbf{X}_k$ and solve the following: }			
 			\State {$\beta_{k+1}^*(\lambda_2) \gets \min_{\beta}~ ||\mathbf{X_k}\beta-Y||_2^2+\lambda_2||\beta||_1$}
			\State {$k \gets k+1$}
			\EndWhile
			\State {Correct the dimension of $\beta$ by innserting $0$ in other indices.}
			\State \textbf{return} $\beta^*$
			\EndProcedure
		\end{algorithmic}
\end{algorithm}

\begin{algorithm}
		\caption{Modified Four-step Recovery}\label{tab:L2SAT}
		\begin{algorithmic}[1]
			\State \textbf{Input:}
			\State  $\mathbf{Y_{m\times1}}: $ {The vector containing the labels}
			\State  $\mathbf{{\hat{X}_{m\times n}}}: $ {The data matrix containing missing entries}
	
			\State  $\mathbf{\epsilon}: $ {Stopping criterion}
			\State  $\mathbf{\gamma}: $ {Threshold level in sifting covariance matrix}
			\State  $(\mathbf{\alpha},\mathbf{\lambda_1},\mathbf{\lambda_2})$ {Algorithm Parameters}
			\State \textbf{Output:}
		\State  $\beta^*$ : {The reconstructed signal}
			
			\Procedure{Four-step Recovery} {${\mathbf Y},{\mathbf {\hat{X}}},{\epsilon},{\alpha}, \lambda_1, \lambda_2$}
\State {Initialization:} $\mathbf{X_0}\gets \mathbf{0}_{m\times n}, k\gets 0$
		\State  $\mathbf{\beta_{0_{n\times1}}} \gets \mathbf{0}_{n\times 1},\mathbf{\beta_{1_{n\times1}}} \gets \mathbf{\hat{X}^{\dagger}} \times \mathbf{Y}$
		    \While {$||\beta_k-\beta_{k+1}||_2>\epsilon$}
 			\State {fix $\beta_k$ and solve the following: }
 			\State {$\mathbf{X_k} \gets \min_{X_k} ~ ||P_E(\mathbf{X_k}-\mathbf{\hat{X}})||_2^2+\lambda_1||\mathbf{X_k}||_*$}
			\State {fix $\mathbf{X}_k$ and solve the following: }			
 			\State {$\beta_{k+1}^*(\lambda_2) \gets \min_{\beta} ~ ||\mathbf{X_k}\beta-Y||_2^2+\lambda_2||\beta||_1$}
			\State {$k \gets k+1$ }			
			\EndWhile
			 			\State {$S \gets \{i: \beta_i \neq 0\}$, $s \gets |S|$ }
			 		\State {Normalize columns of $\mathbf{X_k}$, and define $C\gets \mathbf{X}^T\mathbf{X}$}.
			\State {Confine $\mathbf{\hat{X}}$ on supporting columns in $S$ and add the columns which have correlation larger than $\gamma$ to elements in $S$ based on the Empirical Covariance Matrix ($C$), and denote it with $\mathbf{\hat{X}_{\hat{S}}}$, $\hat{s}\gets|\hat{S}|$. Choose $\alpha<<\epsilon$.}
			\State {Initialization:} $\mathbf{X_0}\gets \mathbf{0}_{m\times s},  k\gets 0$
					\State  $\mathbf{\beta_{0_{\hat{s}\times1}}} \gets \mathbf{0}_{\hat{s}\times 1},\mathbf{\beta_{1_{\hat{s}\times1}}} \gets \mathbf{\hat{X}_{\hat{S}}^{\dagger}} \times \mathbf{Y}$
		    \While {$||\beta_k-\beta_{k+1}||_2>\alpha$}
 			\State {fix $\beta_k$ and solve the following: }
 			\State {$\mathbf{X_k} \gets \min_{X_k} ~ ||P_E(\mathbf{X_k}-\mathbf{\hat{X}_{\hat{S}}})||_2^2+\lambda_1||\mathbf{X_k}||_*$}
			\State {fix $\mathbf{X}_k$ and solve the following: }			
 			\State {$\beta_{k+1}^*(\lambda_2) \gets \min_{\beta}~ ||\mathbf{X_k}\beta-Y||_2^2+\lambda_2||\beta||_1$}
			\State {$k \gets k+1$}
			\EndWhile
			\State {Correct the dimension of $\beta$ by innserting $0$ in other indices.}
			\State \textbf{return} $\beta^*$
			\EndProcedure
		\end{algorithmic}
\end{algorithm}
\begin{figure}
	\centering
	\includegraphics[width=1\linewidth]{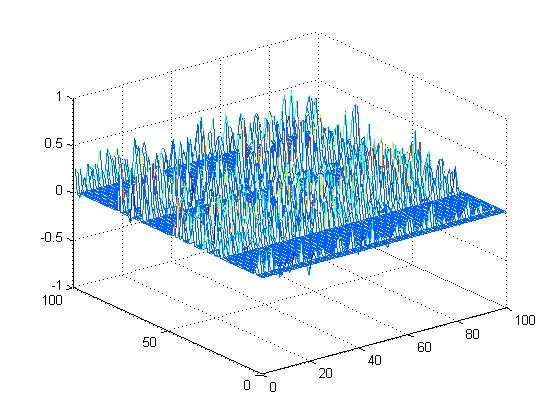}
	\caption{Empirical Covariance estimation for Random Gaussian Data.}\label{fig1}
\end{figure}
\begin{figure}
	\centering
	\includegraphics[width=1\linewidth]{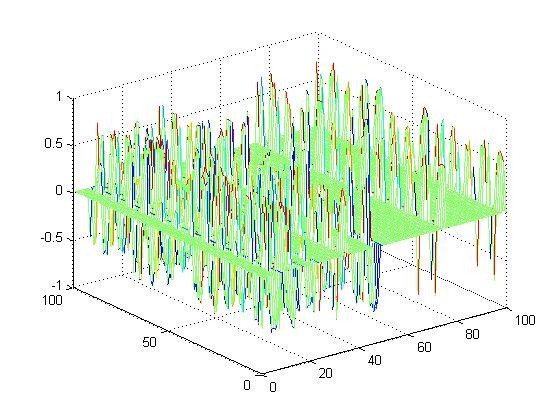}
	\caption{Empirical Covariance estimation for MATLAB Hospital Data1.}\label{fig2}
\end{figure}
\begin{figure}
	\centering
	\includegraphics[width=1\linewidth]{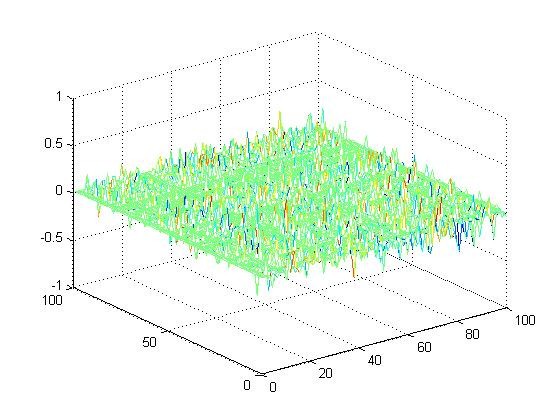}
	\caption{Empirical Covariance estimation for MATLAB Hospital Data2.}\label{fig3}
\end{figure}
\begin{figure}
	\centering
	\includegraphics[width=1\linewidth]{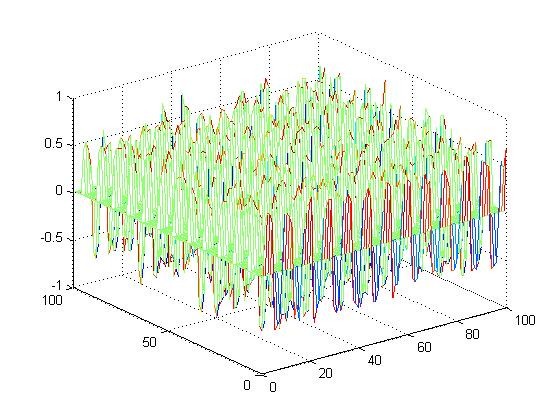}
	\caption{Empirical Covariance estimation for MATLAB Stock Data.}\label{fig4}
\end{figure}
\begin{figure}
	\centering
	\includegraphics[width=1\linewidth]{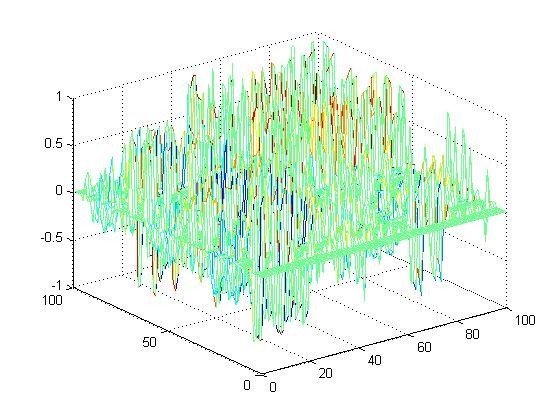}
	\caption{Empirical Covariance estimation for Batch1 Data.}\label{fig5}
\end{figure}
\begin{figure}
	\centering
	\includegraphics[width=1\linewidth]{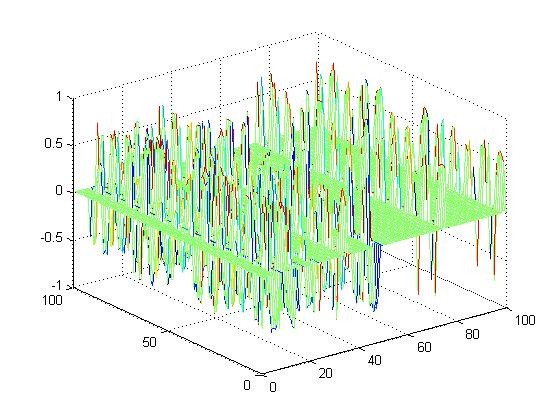}
	\caption{Empirical Covariance estimation for Batch2 Data.}\label{fig6}
\end{figure}
\begin{center}
\pagebreak	
\centering
	\begin{table}[htb]
\caption {Comparison of the RMSE and runtime (in Seconds) for the two methods on various datasets} \label{table1}

	\begin{tabular}{ |p{1.3cm}|p{1.3cm}|p{1.3cm}|p{1.3cm}|p{1.3cm}| }

		\hline
		Datasets & Parameter & Including extra features & Without extra features & Relative variation  \\
		\hline
		\multirow{2}{3cm}{Hospital1} & RMSE & 0.07 & 0.104 & 32\%\\
		\multirow{2}{3cm}{} & runtime & 67.36 & 48.51 & 38.8\%\\
		\hline
				\multirow{2}{3cm}{Hospital2} & RMSE & 0.036 & 0.049 & 26\%\\
				\multirow{2}{3cm}{} & runtime & 62.56 & 55.75 & 12\%\\
\hline
\multirow{2}{3cm}{Stock} & RMSE & 7.24 & 8.05 & 10\%\\
\multirow{2}{3cm}{} & runtime & 12.70 & 11.10 & 14\%\\
\hline
\multirow{2}{3cm}{Batch2} & RMSE & 0.00061 & 0.00073 & 16\%\\
\multirow{2}{3cm}{} & runtime & 31.14 & 26.11 & 19\%\\
\hline
\multirow{2}{3cm}{Batch1} & RMSE & 0.036 & 0.057 & 36\%\\
\multirow{2}{3cm}{} & runtime & 21.12 & 18.53 & 14\%\\
\hline
\multirow{2}{3cm}{Batch1} & RMSE & 0.0016 & 0.002 & 20\%\\
\multirow{2}{3cm}{} & runtime & 26.38 & 19.43 & 36\%\\
\hline		
	\end{tabular}
		\end{table}
\end{center}

\section{Simulation Results}
In this Section, we investigate the simulation results. We notice that in Table \ref{table1} we are reducing Root Mean Squared Error (RMSE) at the cost of more time complexity. In fact, the column with the title "only added features" represents the results of algorithm in \cite{esmaeili2016fast}. The previous column is related to the work we are presenting in this paper. For instance, we can consider the Hospital2 data in MATLAB datasets, and notice that the RMSE in the scenario which includes correlated variables is reduced to $0.036$, while in the ordinary case it is $0.049$. The RMSE is improved for $26\%$. However, the time complexity is increased $12\%$ as a trade-off. As another case to illustrate, we can find in Table \ref{table1} that our modification leads to reducing RMSE from $0.104$ to $0.07$ for Hospital1 Data. Again, the runtime is slightly increased from $48$ Sec's to $67.36$ Sec's and that is the cost of enhancing accuracy. Actually, these two are nearly comparable. The simulations are carried out on Intel(R) core(TM) i7-2670QM CPU @ 2.20 GHz. We notice that the outperformance of our new methodology in enhancing the RMSE error is best observed in the cases where the covariance matrices have non-smooth plots (highly varying) as in figures \ref{fig2}, \ref{fig5}, and \ref{fig6}. The dataset in figures \ref{fig2}, \ref{fig3}, and \ref{fig4} are MATLAB datasets. The data used for figures \ref{fig5}, and \ref{fig6} are derived from \cite{irvine}. We can see two points here. First, we could take advantage much more from the covariance matrices to improve the performance. In addition, it is worth noting that our algorithm could be installed above methods provided in \cite{ganti2015sparse}, and \cite{chan2016temporal} to accelarate their efficacy on big data or highly-featured datasets. 
\section{conclusion}
We introduced a new idea for enhancing prediction accuracy for sparse recovery in linear models with missing data using the concept of nurturing the support of recovered sparse signals by forming the empirical covariance matrix. The line of reasoning we took advantage of is based on the fact that the variables which are not selected in the sparse recovery procedure but are highly correlated to the selected ones, may be helpful in completion of the initial matrix with missing entries even if they do not contribute effectivey to formation of the support of linear models. The time complexity is reduced effectively since we are confining our matrix to the columns relating to the selected features. After we confine it to the nurtured support, we enhance the prediction accuracy at the cost of slightly increased computational complexity. The complexity is still far less than the complexity of algorithms which learn the entire matrix in the literature. Simulation results have shown the effectiveness of our method in enhancing prediction accuracy is more pronounced when the covariance matrix shows a varying behavior.
\bibliographystyle{IEEEtran}
\bibliography{reffff.bib}

\begin{thebibliography}{10}
\providecommand{\url}[1]{#1}
\csname url@samestyle\endcsname
\providecommand{\newblock}{\relax}
\providecommand{\bibinfo}[2]{#2}
\providecommand{\BIBentrySTDinterwordspacing}{\spaceskip=0pt\relax}
\providecommand{\BIBentryALTinterwordstretchfactor}{4}
\providecommand{\BIBentryALTinterwordspacing}{\spaceskip=\fontdimen2\font plus
\BIBentryALTinterwordstretchfactor\fontdimen3\font minus
  \fontdimen4\font\relax}
\providecommand{\BIBforeignlanguage}[2]{{%
\expandafter\ifx\csname l@#1\endcsname\relax
\typeout{** WARNING: IEEEtran.bst: No hyphenation pattern has been}%
\typeout{** loaded for the language `#1'. Using the pattern for}%
\typeout{** the default language instead.}%
\else
\language=\csname l@#1\endcsname
\fi
#2}}
\providecommand{\BIBdecl}{\relax}
\BIBdecl

\bibitem{esmaeili2016fast}
A.~Esmaeili, A.~Amini, and F.~Marvasti, ``Fast methods for recovering sparse
  parameters in linear low rank models,'' \emph{arXiv preprint
  arXiv:1606.08009}, 2016.

\bibitem{5466511}
R.~H. Keshavan, A.~Montanari, and S.~Oh, ``Matrix completion from a few
  entries,'' \emph{IEEE Transactions on Information Theory}, vol.~56, no.~6,
  pp. 2980--2998, June 2010.

\bibitem{candes2009exact}
E.~J. Cand{\`e}s and B.~Recht, ``Exact matrix completion via convex
  optimization,'' \emph{Foundations of Computational mathematics}, vol.~9,
  no.~6, pp. 717--772, 2009.

\bibitem{candes2006compressive}
E.~J. Cand{\`e}s \emph{et~al.}, ``Compressive sampling,'' in \emph{Proceedings
  of the international congress of mathematicians}, vol.~3.\hskip 1em plus
  0.5em minus 0.4em\relax Madrid, Spain, 2006, pp. 1433--1452.

\bibitem{tropp2007signal}
J.~A. Tropp and A.~C. Gilbert, ``Signal recovery from random measurements via
  orthogonal matching pursuit,'' \emph{IEEE Transactions on information
  theory}, vol.~53, no.~12, pp. 4655--4666, 2007.

\bibitem{marvasti2012unified}
F.~Marvasti, A.~Amini, F.~Haddadi, M.~Soltanolkotabi, B.~H. Khalaj,
  A.~Aldroubi, S.~Sanei, and J.~Chambers, ``A unified approach to sparse signal
  processing,'' \emph{EURASIP journal on advances in signal processing}, vol.
  2012, no.~1, p.~1, 2012.

\bibitem{4663911}
H.~Mohimani, M.~Babaie-Zadeh, and C.~Jutten, ``A fast approach for overcomplete
  sparse decomposition based on smoothed $\ell ^{0}$ norm,'' \emph{IEEE
  Transactions on Signal Processing}, vol.~57, no.~1, pp. 289--301, Jan 2009.

\bibitem{esmaeili2016iterative}
A.~Esmaeili, E.~Asadi, and F.~Marvasti, ``Iterative null-space projection
  method with adaptive thresholding in sparse signal recovery and matrix
  completion,'' \emph{arXiv preprint arXiv:1610.00287}, 2016.

\bibitem{doi:10.1137/100781894}
\BIBentryALTinterwordspacing
M.~Tao and X.~Yuan, ``Recovering low-rank and sparse components of matrices
  from incomplete and noisy observations,'' \emph{SIAM Journal on
  Optimization}, vol.~21, no.~1, pp. 57--81, 2011. [Online]. Available:
  \url{http://dx.doi.org/10.1137/100781894}
\BIBentrySTDinterwordspacing

\bibitem{ganti2015sparse}
R.~Ganti and R.~M. Willett, ``Sparse linear regression with missing data,''
  \emph{arXiv preprint arXiv:1503.08348}, 2015.

\bibitem{chan2016temporal}
F.~K.~S. Chan, A.~J. Ma, P.~C. Yuen, T.~C.-F. Yip, Y.-K. Tse, V.~W.-S. Wong,
  and G.~L.-H. Wong, ``Temporal matrix completion with locally linear latent
  factors for medical applications,'' \emph{arXiv preprint arXiv:1611.00800},
  2016.

\bibitem{goldberg2010transduction}
A.~Goldberg, B.~Recht, J.~Xu, R.~Nowak, and X.~Zhu, ``Transduction with matrix
  completion: Three birds with one stone,'' in \emph{Advances in neural
  information processing systems}, 2010, pp. 757--765.

\bibitem{irvine}
\BIBentryALTinterwordspacing
 [Online]. Available:
  \url{https://archive.ics.uci.edu/ml/machine-learning-databases/00224/}
\BIBentrySTDinterwordspacing

\end{thebibliography}
\end{document}